# Fast and accurate classification of echocardiograms using deep learning

Ali Madani[1], Ramy Arnaout MD DPhil[2], Mohammad Mofrad PhD[1], Rima Arnaout MD[3]*

[1] 208A Stanley Hall Room 1762
Molecular Biophysics and Integrated Bioimaging Division, Lawrence Berkeley National Lab
California Institute for Quantitative Biosciences (QB3)
University of California at Berkeley
Berkeley, CA 94720
madani@berkeley.edu
mofrad@berkeley.edu

[2] 330 Brookline Avenue Dana 615
Beth Israel Deaconess Medical Center
Boston, MA 02215
rarnaout@bidmc.harvard.edu

[3] 555 Mission Bay Blvd South Rm 484
Cardiovascular Research Institute
University of California, San Francisco, 94143
rima.arnaout@ucsf.edu
* Corresponding Author

**Abstract**

**Background**—Echocardiography is essential to modern cardiology. However, human interpretation limits accurate and standardized high-throughput analysis, limiting echocardiography from reaching its full clinical and research potential for precision medicine. Deep learning is a cutting-edge machine-learning technique that has been useful in analyzing medical images but has not yet been widely applied to echocardiography, partly due to the complexity of echocardiograms' multi-view, multi-modality format. The essential first step toward comprehensive computer-assisted echocardiographic interpretation is determining whether computers can learn to recognize standard views.

**Goals**—To test whether deep learning can accurately classify echocardiogram views.

**Methods**—We anonymized 834,267 transthoracic echocardiogram (TTE) images from 267 patients (20-96 years, 51% female, 26% obese) seen between 2000 and 2017 and labeled them according to standard views. Images covered a range of real-world acquisitional and clinical variation. We built a multilayer convolutional neural network and used supervised learning to simultaneously classify 15 standard views. Eighty percent of data used was randomly chosen for training and 20% reserved for validation and testing on never-seen echocardiograms.

**Results**—Using multiple images from each clip, the model classified among 12 video views with 97.8% overall test accuracy without overfitting. Even on single low-resolution images, test accuracy among 15 views was 91.7% versus 70.2-83.5% for board-certified echocardiographers. Confusional matrices, occlusion experiments, and saliency mapping showed that the model finds recognizable similarities among related views and classifies using clinically relevant image features.

**Conclusions**—Deep neural networks can classify essential echocardiographic views simultaneously and with high accuracy. Our results provide a foundation for more complex deep-learning-assisted echocardiographic interpretation.

**Introduction**

Imaging is a critical part of medical diagnosis. Interpreting medical images typically requires extensive training and is a complex and time-intensive process. Deep learning, specifically using convolutional neural networks (CNNs), is a cutting-edge machine learning technique that has proven "unreasonably" (1) successful at learning patterns in images and has shown great promise helping experts with image-based diagnosis in radiology, pathology, dermatology, and other fields, for example in detecting the boundaries of organs in computed tomography and magnetic-resonance images, flagging suspicious regions on tissue biopsies, and classifying photographs of benign vs. malignant skin lesions (2-4). However, deep learning has not yet been widely applied to echocardiography, a noninvasive, relatively inexpensive, radiation-free imaging modality indispensable to modern cardiology (5).

An echocardiogram consists of scores of video clips, still images, and Doppler recordings measured from over a dozen different acquisition angles, called "views." The majority of the information is represented as video clips; only pulsed-wave Doppler (PW), continuous-wave Doppler (CW), and m-mode recordings are represented exclusively as single images. Determining the view is the essential first step in interpreting an echocardiogram (6). This step is non-trivial, not least because several views differ only subtly from each other. In principle, a CNN can be trained to classify views, requiring only a training set of labeled images from which to learn; given a new image, a well-trained model should then be able determine the view from a new image instantaneously. The simplicity of training in deep learning represents a significant advantage over earlier machine-learning methods, which have sometimes been applied to echocardiography but often require time-consuming and operator-dependent manual selection and annotation of features, such as manually tracing the outline of the heart, in each of a large number of training images (7-11).

To assist echocardiographers and improve use of echocardiography for precision medicine, we tested whether supervised deep learning with CNNs can be used to automatically classify views without requiring prior manual feature selection. We report a model that achieves almost 98 percent overall test accuracy based on a variety of video and still-image view-classification tasks.

To achieve translational impact in medicine, novel computational models must not just achieve high accuracy but must also address clinical relevance. We did this in two main ways. First, we used randomly selected, real-world echocardiograms to train our model, including a variety of patient variables, echocardiographic indications and pathologies, technical qualities, and multiple vendors to ensure that our deep learning model would be clinically relevant. Second, deep-learning models are sometimes considered "black boxes" because their internal workings are at first glance obscure. To address this issue, we use several methods look inside our model to show that classification depends on human-recognizable clinical features within images.

Taken together, these results suggest that our approach may be useful in helping echocardiographers improve their accuracy, efficiency, and workflow and may provide a foundation for high-throughput analysis of echocardiographic data.

**Methods**

*Dataset.* All datasets were obtained and de-identified in compliance with the Institutional Review Board at the University of California, San Francisco (UCSF). Two-hundred sixty-seven echocardiographic studies performed between 2000 and 2017 were selected at random from UCSF's clinical database. These studies included men and women (49.4 and 50.6 percent, respectively) ages 20-96 (median age, 56; mode, 63) with a range of body types (25.8 percent obese) which can affect technical quality of TTE, and included indications and pathologies that are representative of the uses of echocardiography in current clinical practice (**Table 1**). Studies were carried out using echocardiographic equipment from several manufacturers (e.g., GE, Phillips, Siemens).

*Data processing.* DICOM-formatted echocardiogram videos and still images were stripped of identifying metadata, labeled by view, split into constituent frames, anonymized by zeroing out all pixels that contained identifying information, and converted into standardized 60x80-pixel monochrome images, resulting in 834,267 images. Fifteen views were selected for multi-category classification, covering the majority used in the field. Views classified included parasternal long axis, right ventricular inflow, basal short axis (aortic valve level), short axis at mid (papillary muscle) or mitral level, apical four-chamber, apical five chamber, apical two chamber, apical three chamber (apical long axis), subcostal four-chamber, subcostal inferior vena cava (IVC), subcostal abdominal aorta, suprasternal aortic arch, pulsed-wave Doppler, continuous-wave Doppler, and m-mode. For each view, we included images with a range of natural echocardiographic variation, such as differences in zoom, depth, focus, sector width, gain, chroma map, systole/diastole, angulation, image quality, and use of 3D, color Doppler, dual mode, strain, and left-ventricular (LV) contrast, to capture the range of variation normally seen by echocardiographers.

A subset of 223,787 images from 15 views were split into training, validation, and test datasets in approximately an 80:10:10 ratio. Each dataset contained images from separate echocardiographic studies, to maintain sample independence. The number of images in training, validation, and test datasets were 180,294, 21,747, and 21,746 images, respectively. The validation dataset was used for model selection and parameter fine-tuning. The test dataset was used for performance evaluation of the final trained and validated model. For training, 256-shade greyscale pixel values were scaled from [0,255] to [0,1] and the mean over the training data was subtracted from each dataset, as is standard in image-recognition tasks. Also as per standard practice, data was augmented at run-time by randomly applying rotations of up to 10 degrees, width and height shifts of up to a tenth of total length, zooms of up to 0.08, shears of up to 0.03, and vertical/horizontal flips. Training and validation datasets in which view labels were randomized were used as a negative control.

*Model architecture and training.* Our neural network architecture was designed in Python using the Tensorflow, Theano, and Keras packages, drawing inspiration from the VGG16 network, which won the Imagenet challenge in 2014 (12-15). Our model utilized a series of small 3x3 convolutional filters connected with max-pooling layers over 2x2 windows. Dropout was utilized in training for both the convolutional and fully connected layers to prevent overfitting. In addition to dropout for regularization, batch normalization was used before neuron activations which led to faster training and increased accuracy. Activation functions were mainly rectified linear units (ReLU) with the exception of the softmax classifier layer. Training was performed over 45 epochs using an adaptive learning-rate decay for RMSprop optimization. $k$-fold cross-validation was used to randomly vary which images were in the training and validation sets, to make use of all available data for training and to select the optimal weights at each epoch. Batches of 64 samples at a time were used for gradient calculation. Convergence plots of training and validation accuracy by epoch confirmed that the model was not overfitting. The training method was robust, with three separate trainings of the 223,787 images resulting in overall test accuracies above 97 percent. Training was performed on Amazon's EC2 platform with a GPU instance g2.2large.

*Model evaluation.* Several metrics were used over the test dataset for performance evaluation. Overall accuracy was calculated as the number of correctly classified images as a fraction of the total number of images. Average accuracy was calculated as the average over all views of per-view accuracy. F-score was calculated in standard fashion as twice the harmonic mean of precision (positive predictive value) and recall (sensitivity). Receiver operator characteristic (ROC) curves were plotted in the standard way as true-positive fraction (y-axis) against false-positive fraction (x-axis) and the associated area under curve (AUC) was calculated. Confusion matrices were calculated and plotted as heat maps to visualize performance of multi-view classifiers and their associated errors. Single test images were classified according to the view with the highest probability. Test videos were classified by simple majority vote on multiple images from a given video.

The basis for the model's classification decisions was explored using t-distributed stochastic neighbor embedding (t-SNE) dimensionality reduction (16) of raw pixels and the last fully-connected layer output for each sample. Occlusion experiments were performed by masking test images with bounding boxes of different shapes, then submitting them to the model for label prediction. Saliency maps were created using guided backpropogation, which keeps the model weights fixed and computes the gradient of the model's output for a given image.

*Comparison to human experts.* Echocardiogram test-image classification by board-certified echocardiographers was approved by the UCSF Human Research Protection Program and Institutional Review Board. Each board-certified echocardiographer was given a randomly selected subset of 1,500 images, 100 of each view, drawn from the test set given to the model, and performance compared using the relevant metrics above.

**Results**

*Deep learning achieves expert-level view classification*. We designed and trained a CNN (**Figure 1**) to recognize 15 different standard echocardiographic views, 12 from b-mode and three from PW, CW, and m-mode (**Figure 2**), using a training and validation set of over 200,000 images and a test set of over 20,000 images. These images covered a range of natural echocardiographic variation, both with respect to patient variables (**Table 1**) and to differences in zoom, depth, focus, sector width, gain, chroma map, systole/diastole, angulation, image quality, and use of 3D, color Doppler, dual mode, strain, and LV contrast (**Figure 3**). Clustering analyses showed that the neural network could sort heterogeneous input images into groups according to view (**Figure 4**).

The model achieved an average overall test accuracy of 97.8 percent on videos (F-score 0.964 ± s.d. 0.035) and 100 percent accuracy on seven of the 12 video views (**Figure 5A**). CW, PW, and m-mode categories, which always appear in echocardiograms as still images, had 98, 83, and 99 percent accuracies, respectively (**Figure 5B**).

On single still images drawn from all 15 views, the model achieved an average overall accuracy of 91.7 percent (F-score 0.904 ± s.d. 0.058) (**Figure 5B**), compared to 76.9 percent (range, 70.2-83.5; n=2 subjects) for board-certified echocardiograpers (**Figure 5C**). AUCs for still-image model prediction by view category ranged from 0.985-1.00 (mean 0.996; **Figure 5F**). For the 8.3 percent of test images that the model misclassified, its second-best guess—the view with the second-highest probability—was the correct one in 67.0 percent of cases (5.3 percent of test images; **Figure 5E**). Therefore, 97.3 percent of test still-images were classified correctly when considering the model's top two guesses.

Accuracy was highest for views with more training data (e.g. apical 4-chamber) and views that are most visually distinct (e.g. m-mode). Accuracy was lowest for views that were clinically similar to other views, such as apical three-chamber (which can be confused for apical two-chamber) vs. apical four-chamber (vs. apical five-chamber), or views in which multiple view-defining structures can be seen in the same image, such as subcostal IVC vs. subcostal four-chamber. As expected, training on randomly labeled still images achieved an accuracy (6.9 percent) commensurate with random guessing (6.7 percent, the probability of guessing the correct one out of 15 views purely by chance).

*Model classification is based on clinically relevant features.* To understand whether classification is based on clinically relevant features, such as heart chambers and valves, or on confounding or statistical features that might be clearer to a machine than a human, such as fiducial marks or fraction of white pixels, we performed occlusion experiments by measuring prediction performance on test images on which we masked clinically relevant features with different shapes. Overall test accuracy fell with masking of the heart but not other parts of the image, consistent with this region being important to the model (**Figure 6A**). In addition, saliency mapping, which identifies the input pixels that are most important to the model's assignment of a particular classification, revealed that structures that would be important to defining the view to a human expert were the ones that contributed most to the model's classification (**Figure 6B**).

**Discussion**

View classification is the essential first step in the interpretation of echocardiograms. Previous attempts to use machine learning to assist with view classification required laborious manual annotation, failed to distinguish among more than a few views at a time, used only "textbook-quality" images, exhibited low accuracy, or were tied to a specific equipment vendor, limitations unsuitable for general practice (7-11, 17). In contrast, ours is the first report to our knowledge of a single, vendor-agnostic, deep-learning model that correctly classifies all types of echocardiogram recordings (b-mode, m-mode, and Doppler; still images and videos) from all acquisition points relevant to a full standard transthoracic echocardiogram (parasternal, apical, subcostal, and suprasternal), at accuracies that exceed those of board-certified echocardiographers. Furthermore, the echocardiograms used in this study were drawn randomly from real echocardiograms acquired for clinical purposes, from patients with a range of ages, sizes, and hemodynamics; for a range of indications; and including a range of pathologies such as low left ventricular ejection fraction, left ventricular hypertrophy, valve disease, pulmonary hypertension, pericardial effusion. Training data also included the natural variation in echocardiographic acquisition of each view, including variations in technical quality. By avoiding limited or idealized training subsets, our model is broadly applicable to clinical practice.

Because deep networks like CNNs usually include large numbers of (highly correlated) parameters (which describe the weights of connections among the nodes in the network), it is usually difficult to understand a model's decision-making by simple inspection. For life-or-death decisions such as in medicine or self-driving cars, this issue can breed suspicion and has legal ramifications that can slow adoption. Occlusion testing and saliency mapping help address these concerns by getting inside the black box. In our model, these techniques show that classification depends on the same features that echocardiographers use to reach their conclusions. For example, the maps shown in Figure 7B for a short-axis-mid view and a suprasternal-aorta view, respectively, each trace the basic outlines of their corresponding input view. In the future, applying these approaches to intermediate layers may prove interesting to more precisely define the similari-ties, or differences, in how humans and models move from features to conclusions. For now, it is reassuring that our model considers the same features that human experts do in classifying views.

This similarity also explains the occasional misclassifications of single images, which most often involved views that can look similar to human eyes (**Figure 2 E&F, G&H, J&K; Figure 5**). These include adjacent views in echocardiographic acquisition, where a slight difference in the angle of the sonographer's wrist can change the view, resulting in confusion of an apical three-chamber view for an apical two-chamber view or an apical five-chamber for apical four-chamber; as well as views in which two view-defining structures may be seen in the same image, such as the IVC seen in a subcostal four-chamber view. A low-velocity PW signal can look similar to a faint CW signal. In fact, the only misclassification made by our model without an obvious explanation of this sort was that of the right ventricular inflow view for short-axis basal; of note, the right ventricular inflow view was also very challenging for human echocardiographers to distinguish (with 51-57 percent accuracy). We note in the confusion matrices that misclassification of certain views for one another was non-symmetrical; for example, PW images were confused with CW, but CW images were almost never mistaken for PW (**Figure 5B**). In this case, as mentioned above, this asymmetry makes clinical sense, however, more training and test data can be used to explore this phenomenon further and refine accuracies for these categories. Because classification of videos is based on multiple images, and error decays exponentially with the number of images, misclassification of videos was very rare (~2 percent; **Figure 5D**). We also noted that the model's confidence in its choice (the probability assigned to a view classification for a particular image) affected performance; where confidence was higher, accuracy was also higher **(Figure 7)**. Therefore, communicating the model's confidence for each classification should further benefit users.

Finally, our approach had two unexpected advantages related to efficiency, practicability, and cost-effectiveness. First was the perhaps surprising effectiveness of a simple majority vote in classification of videos. Video analysis can be a complex undertaking that involves non-trivial tasks such as frame-to-frame color variation and object tracking. We have demonstrated that view classification, at least, can be done much more efficiently and cost-effectively, reducing coding and training time. Moving beyond view classification, it will be interesting to see what other clinically actionable information can be extracted from (collections of) still images. Second, in removing color and in standardizing the sizes and shapes of videos and still images for training, we discovered that we could downsample—i.e., shrink—images appreciably without losing accuracy. This allowed for a 25-491-fold savings in file size (vs. 300-by-400- to 1024-by-768-pixel images; **Figure 8**), and corresponding gains in the cost and speed of training and in classification. We note the potential implications for telemedicine, including in resource-poor regions of the United States and elsewhere, of requiring storage and transmission of smaller files (though decentralized use of the model can also come through transmission of the model, which is a small file).

Echocardiography is essential to diagnosis and management for virtually every cardiac disease. In this study, we have demonstrated the application of deep learning to echocardiography view classification that classified 15 major TTE views with expert-level quality. We purposely used a training set that reflected a wide range of clinical and physiological variation, demonstrating ap-plicability to real-world data. We found that our model uses some of the same features in echocardiograms that human experts use to make their decisions. Looking forward, our model can be expanded to classify additional sub-categories of echocardiographic view, as well as diseases, work that has foundational utility for research, for clinical practice, and for training the next generation of echocardiographers.

**Acknowledgements**

We thank the board-certified echocardiographers who scored images as human experts.

**Table 1. Echocardiogram characteristics.**

|  | Mean | SD |
|---|---|---|
| Age (years) | 56.1 | 16.6 |
| Height (cm) | 169.5 | 11.6 |
| Weight (kg) | 77 | 20.5 |
| Systolic BP (mmHg) | 127.1 | 19.3 |
| Diastolic BP (mmHg) | 69.8 | 12.7 |
| MAP (mmHg) | 88.9 | 13.4 |
| BSA (m$^2$) | 1.87 | 0.27 |
| BMI (kg/m$^2$) | 26.6 | 6.1 |
|  | **Percent** | **N** |
| Female | 50.6 | 135 |
| Male | 49.4 | 132 |
| Obese | 25.8 | 69 |
| LVEF <55% | 21.7 | 58 |
| LVEDVI > normal (sex adjusted) | 16.9 | 45 |
| LVMI > normal (sex adjusted) | 32.5 | 87 |
| RVSP > 40 mmHg | 8.4 | 22 |
| TAPSE < 1.6 cm | 4.8 | 13 |
| **Indication** | **Percent** | **N** |
| Heart Failure/ Cardiomyopathy | 24.0 | 64 |
| Arrhythmia | 11.6 | 31 |
| Chemotherapy | 10.9 | 29 |
| Valve Disease | 10.5 | 28 |
| Preoperative exam | 7.9 | 21 |
| Dyspnea | 6.4 | 17 |
| Coronary Artery Disease | 6.0 | 16 |
| Stroke | 6.0 | 16 |
| Syncope | 5.2 | 14 |
| Rule out Endocarditis | 4.9 | 13 |
| Pulmonary HTN | 4.5 | 12 |
| Hypertension | 3.7 | 10 |
| Pericardial Effusion | 3.4 | 9 |
| Murmur | 3.0 | 8 |
| Palpitations | 3.0 | 8 |
| Aortic Aneurysm | 2.6 | 7 |
| Congenital Heart Disease | 2.6 | 7 |
| Lung Disease | 1.9 | 5 |
| Edema | 1.5 | 4 |
| Hypotension | 1.5 | 4 |
| Cardiac Arrest | 0.4 | 1 |
| Heart Transplant | 0.4 | 1 |

BP = blood pressure. MAP = mean arterial pressure. BSA = body surface area. BMI = body mass index. HTN = hypertension. LVEF = left ventricular ejection Fraction. LVEDVI = left ventricular end-diastolic volume index (ml/m$^2$). LVMI = left ventricular mass index (g/m$^2$). RVSP = right ventricular systolic pressure. TAPSE = tricuspid annular plane systolic excursion.

**Figure Legends**

**Figure 1. Convolutional neural net architecture for image classification.** The neural network algorithm used for classification included six convolutional layers and two fully-connected layers of 1028 and 512 nodes, respectively. The softmax classifier (pink circles) consisted of up to 15 nodes, depending on the classification task at hand. Conv = convolutional layer; Max Pool = max pooling layer; FC = fully connected layer.

**Figure 2. Sample input images.** Views classified included parasternal long axis (A), right ventricular inflow (B), basal short axis (C), short axis at mid or mitral level (D), apical four-chamber (E), apical five chamber (F), apical two chamber (G), apical three chamber/apical long axis (H), subcostal four-chamber (I), subcostal inferior vena cava (J), subcostal/abdominal aorta (K), suprasternal aorta/aortic arch (L), pulsed-wave Doppler (M), continuous-wave Doppler (N), and m-mode (O). Note that these images are the actual size and resolution of input data to the deep learning algorithm.

**Figure 3. Natural variations in input data.** In addition to applying data augmentation algorithms, we included in each category a range of images representing the natural variation seen in real-life echocardiography. The parasternal long-axis view is shown here for example (A). Variations include diastole (A) vs. systole (B), differences in gain or chroma map (C), use of dual-mode acquisition (D), differences in depth (E) and zoom (F), technically challenging images (G), use of 3D acquisition (H), a range of pathologies (seen here in (I), concentric left ventricular hypertrophy and pericardial effusion), and use of color Doppler (J), as well as differences in angulation, sector width, and use of LV contrast.

**Figure 4. Deep learning model simultaneously distinguishes among 15 standard echocardiographic views.** We developed a deep-learning method to classify among standard echocardiographic views, represented here by t-SNE clustering analysis of image classification. On the left, t-SNE clustering of input echocardiogram images. Each image is plotted in 4800-dimensional space according to the number of pixels, and projected to 2-dimensional space for visualization purposes. Different colored dots represent different view classes (see legend in Figure). Prior to neural network analysis, input data does not cluster into clear groups. On the right, data as processed through the last fully-connected layer of the neural network are again represented in 2-dimensional space, showing organization into clusters according to view category. Abbreviations are as follows: a4c, apical 4 chamber; psla, parasternal long axis; saxbasal, short axis basal; a2c, apical 2 chamber; saxmid, short axis mid/mitral; a3c, apical 3 chamber; sub4c, subcostal 4 chamber; a5c, apical 5 chamber; ivc, subcostal ivc; rvinflow, right ventricular inflow; supao, suprasternal aorta/aortic arch; subao, subcostal/abdominal aorta; cw, continuous-wave Doppler; pw, pulsed-wave Doppler; mmode, m-mode recording.

**Figure 5. Echocardiogram view classification by deep learning model.** Confusional matrices showing actual view labels on y-axis, and neural network-predicted view labels on the x-axis by view category for video classification (A) and still-image classification (B) compared with a representative board-certified echocardiographer (C). Reading across true-label rows, the numbers in the boxes represent the percentage of labels predicted for each category. Color intensity corresponds to percentage, see heatmap on far right; the white background indicates zero percent. Categories are clustered according to areas of the most confusion. Rows may not add up to 100

percent due to rounding. (D) Comparison of accuracy by view category for deep-learning-assisted video classification, still-image classification, and still-image classification by a representative echocardiographer. (E) A comparison of percent of images correctly predicted by view category, when considering the model's highest-probability top hit (white boxes) vs. its top two hits (blue boxes). (F) Receiver operating characteristic curves for view categories were very similar, with AUCs ranging from 0.985-1.00 (mean 0.996). Abbreviations are as follows: saxmid, short axis mid/mitral; ivc, subcostal ivc; subao, subcostal/abdominal aorta; supao, suprasternal aorta/aortic arch; saxbasal, short axis basal; rvinflow, right ventricular inflow; a2c, apical 2 chamber; a3c, apical 3 chamber; a4c, apical 4 chamber; a5c, apical 5 chamber; psla, parasternal long axis; sub4c, subcostal 4 chamber.

**Figure 6. Visualization of decision-making by neural network.** (A) Occlusion experiments. All test images (a short axis basal sample image is shown here) were modified with grey masking of different shapes and sizes as shown, and test accuracy predicted for the test set based on each different modification. Masking which covered cardiac structures resulted in the poorest predictions. (B) Saliency maps. The input pixels weighted most heavily in the neural network's classification decision for two example images (left; suprasternal aorta/aortic arch and short axis mid/mitral input examples shown) were calculated and plotted. The most important pixels (right) make an outline of structures clinically relevant to the view shown.

**Figure 7. Confidence for first and second guesses on image classification.** Box plot summarizing probabilities assigned to correct and incorrect images in the test set using the single highest probability to classify the test image. Confidence for correct answers was higher than for in-correct answers. Median, interquartile range for correct (0.999, 0.970-1.00) and incorrect (0.682, 0.525-0.867) answers.

**Figure 8. Example native-resolution image.** Native echocardiographic images ranged from 300x400 to 768x1024 pixels in resolution, and many contained color, either as color Doppler or as different chroma maps to aid in visualization.

**Figure 1. Convolutional neural net architecture for image classification.**

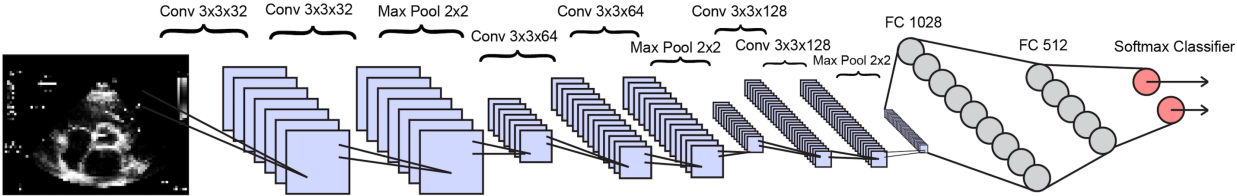

**Figure 2. Sample input images.**

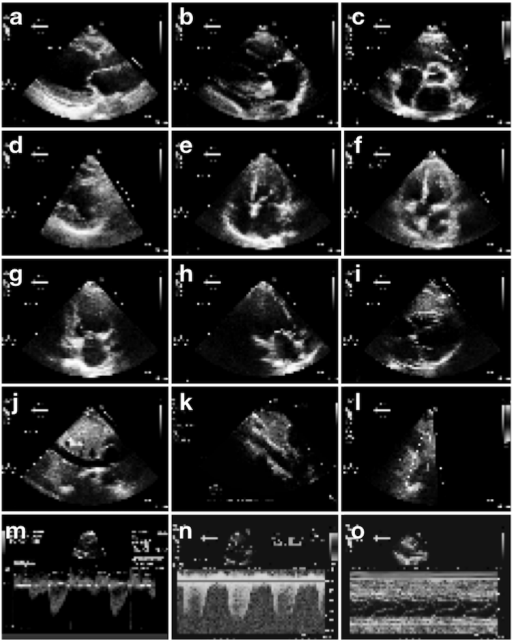

**Figure 3. Natural variations in input data.**

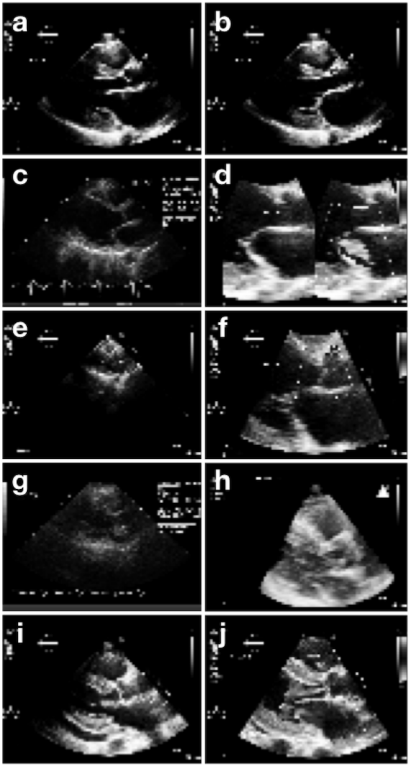

**Figure 4. Deep learning model simultaneously distinguishes among 15 standard echocardiographic views.**

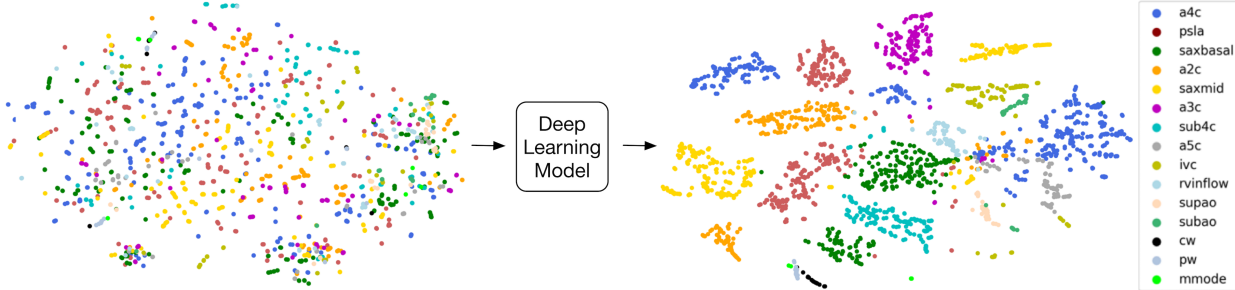

**Figure 5. Echocardiogram view classification by deep learning model.**

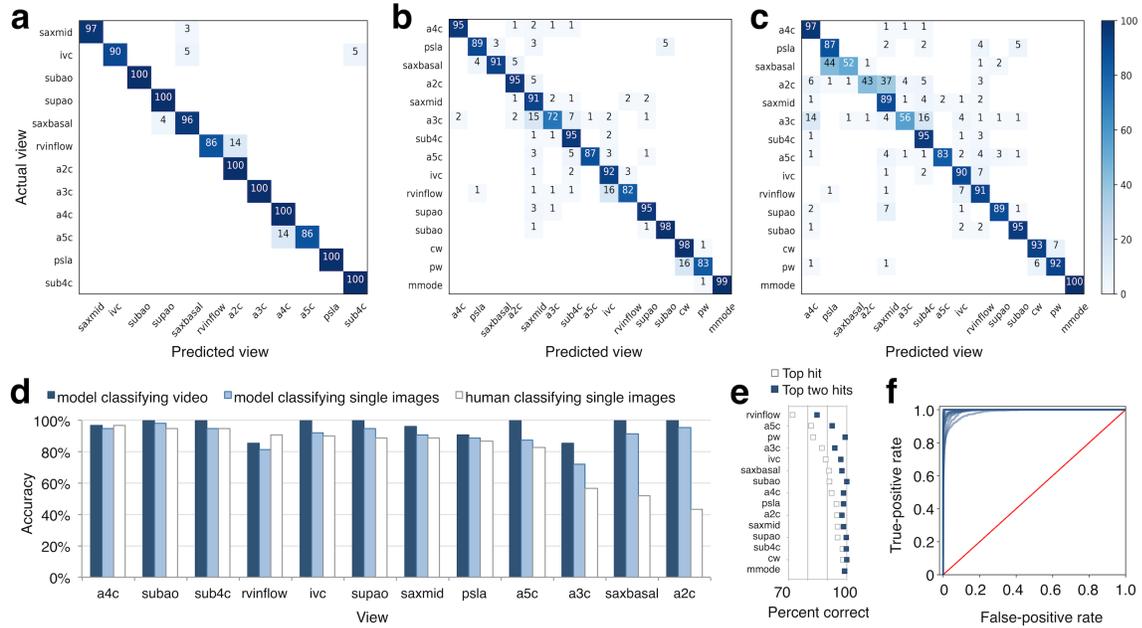

**Figure 6. Visualization of decision-making by neural network.**

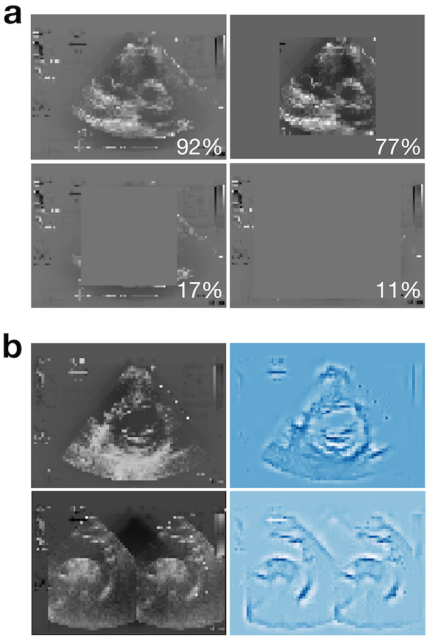

**Figure 7. Confidence for first and second guesses on image classification.**

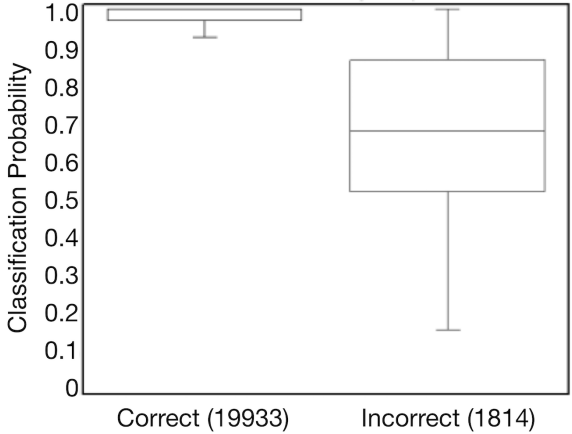

**Figure 8. Example native-resolution image.**

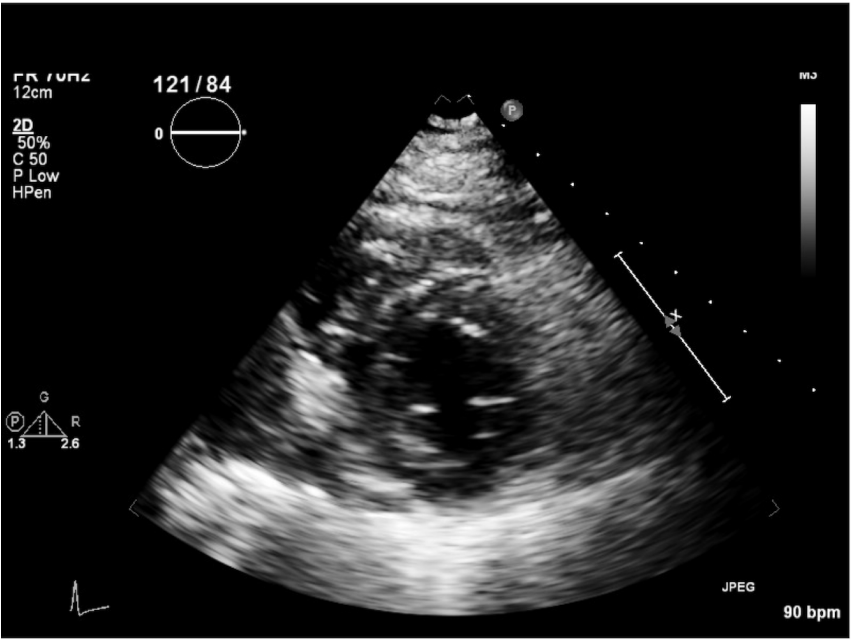